\documentclass[conference]{IEEEtran}
\IEEEoverridecommandlockouts
\usepackage{cite}
\usepackage{amsmath,amssymb,amsfonts}
\usepackage{algorithmic}
\usepackage{graphicx}
\usepackage{textcomp}
\usepackage{xcolor}

\usepackage[ruled,linesnumbered]{algorithm2e}
\usepackage{multirow}
\usepackage{makecell}
\usepackage{enumitem}
\setitemize[1]{noitemsep,partopsep=0pt,parsep=0pt,topsep=0pt, leftmargin=10pt,
rightmargin=0pt}
\def\BibTeX{{\rm B\kern-.05em{\sc i\kern-.025em b}\kern-.08em
    T\kern-.1667em\lower.7ex\hbox{E}\kern-.125emX}}

\begin{document}

\title{Cognition-Mode Aware Variational Representation Learning Framework for Knowledge Tracing}

\author{\IEEEauthorblockN{Moyu Zhang, Xinning Zhu$^{\ast}$ \thanks{*Corresponding author}, Chunhong Zhang, Feng Pan, Wenchen Qian, Hui Zhao}
\IEEEauthorblockA{\textit{Beijing University of Posts and Telecommunications}\\
Beijing, Beijing \\
zhangmoyu@bupt.cn, \{zhuxn, zhangch, Pan\_Feng, wenchen, hzhao\}@bupt.edu.cn}
}

\maketitle

\begin{abstract}
The Knowledge Tracing (KT) task plays a crucial role in personalized learning, and its purpose is to predict student responses based on their historical practice behavior sequence. However, the KT task suffers from data sparsity, which makes it challenging to learn robust representations for students with few practice records and increases the risk of model overfitting. Therefore, in this paper, we propose a Cognition-Mode Aware Variational Representation Learning Framework (CMVF) that can be directly applied to existing KT methods. Our framework uses a probabilistic model to generate a distribution for each student, accounting for uncertainty in those with limited practice records, and estimate the student's distribution via variational inference (VI). In addition, we also introduce a cognition-mode aware multinomial distribution as prior knowledge that constrains the posterior student distributions learning, so as to ensure that students with similar cognition modes have similar distributions, avoiding overwhelming personalization for students with few practice records. At last, extensive experimental results confirm that CMVF can effectively aid existing KT methods in learning more robust student representations. Our code is available at https://github.com/zmy-9/CMVF.
\end{abstract}


In recent decades, numerous computer-assisted learning platforms have emerged to assist students in acquiring knowledge. Recommending suitable problems to students is a crucial function of online platforms, as it prevents students from wasting time practicing questions they already mastered \cite{learn_rec2, mob}. As a result, the Knowledge Tracing (KT) \cite{kt_task, kt_survey} task has received significant attention in recent years, aiming to predict the probability of students answering the target question correctly based on their historical practice sequences.

As a popular data mining application field, KT has produced many excellent models that enhance model prediction accuracy through improving model structures and introducing rich features. The current state-of-the-art KT models basically adopt an \emph{Input $\rightarrow$ Embedding $\rightarrow$ Neural Network $\rightarrow$ Prediction} paradigm, in which the \emph{Embedding} module maps input factors to dense representation vectors, playing a critical role in this paradigm. Thus, exploring how to reasonable representation vectors of input factors, including questions and students, is an active research direction in the KT field. For example, PEBG \cite{pebg} and MF-DAKT \cite{mf_dakt} construct question graphs with question difficulty and relationship information to pre-train question representations, while RKT \cite{rkt} and HGKT \cite{hgkt} introduce question texts to enrich question representations. However, the aforementioned methods focus primarily on question representations, with few works focusing on student representation learning, especially for \emph{Infrequent Students} (students with few practice records). Infrequent students have limited practice records, causing them to have large uncertainties, which lead to low confidence in the model's scoring and makes the existing KT models prone to overfitting.

\begin{figure}[t]
  \centering
  \includegraphics[width=\linewidth]{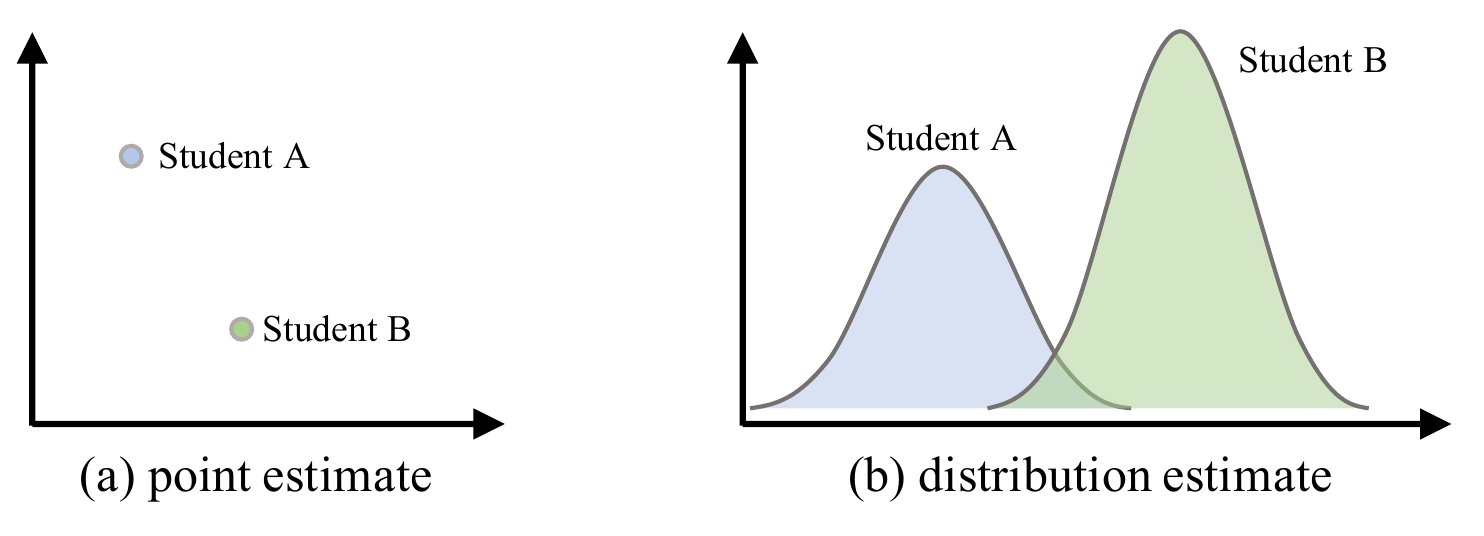}
  \caption{Illustration of point estimation and distribution estimation, where (a) shows that the representation vectors of students A and B can be projected as two points in the embedding space, and the model mainly learns the positions of these two points. On the contrary, (b) shows that students A and B have different distributions, each distribution contains an infinite number of points, representing the uncertainty of the students, and the model needs to learn the parameters of these two distributions.}
  \label{example}
  \vspace{-0.5cm}
\end{figure}

To address student representation overfitting due to data sparsity, KTM \cite{ktm} previously proposed to introduce rich student attribute features. However, these attribute features are often shared by frequent and infrequent students. In the model training stage, due to the influence of data imbalance, that is, frequent student records are significantly more than infrequent student records, attribute features will better reflect the characteristics of frequent students, leading to the individualization of infrequent students being overwhelmed. Later, CL4KT \cite{cl4kt} proposed  to apply data augmentation methods, such as clipping or reordering historical practice behaviors, to improve the robustness of student representations. However, since infrequent learners have limited historical practice, clipping or reordering may lead to information diversity and result in limited information gain or even introduce noise. Furthermore, the above two methods both use point estimation to learn student representations, that is, try to learn a reliable single point for each student in the embedding space, and fail to consider the uncertainty of students, as shown in Figure \ref{example}(a). This results in the model having lower confidence in the prediction scores (i.e., overfitting). Therefore, to effectively model the uncertainty of infrequent students, we propose a general framework called \textbf{C}ognition-\textbf{M}ode Aware \textbf{V}ariational Representation Learning \textbf{F}ramework (CMVF) to generate robust representations for students.

To improve modeling of student uncertainty, we propose to replace point estimate in KT models with distributional estimate, shown in Figure \ref{example}(b). Specifically, we introduce a probabilistic model that learns a distribution for each student to model their uncertainty, calculating the expected value of the distribution as the final prediction score, thereby reducing uncertainty more efficiently. To avoid the computational difficulty of distribution estimate, we propose to use a variational inference-based (VI) approach to build a probabilistic latent variable model \cite{vae}, which enables us to use the reparameterization trick to generate different distributions for each student, thereby facilitating integration with existing deep KT models.

Meanwhile, VI commonly sets a uniform prior distribution for all student distributions, limiting the model's ability to personalize infrequent students. Generally speaking, students with similar characteristics should have similar prior distributions to enable the model to learn similar posterior distributions for them, facilitating sharing of global characteristic information while retaining individuality of infrequent students. Previous KT research shows each student has unique learning characteristics reflected in their practice sequences \cite{dimkt, iekt}, such as different students may acquire knowledge at different speeds in practice, we refer to this characteristic as \emph{cognition mode}. As a result, students with similar cognition modes should have similar representation distributions. To extract students' cognition modes, we propose to utilize the dynamic routing algorithm \cite{capsule} to model their practice sequences, with different \emph{capsules} representing different cognition modes. Based on extracted cognition modes, we design a cognition-mode aware multinomial distribution as a prior during Bayesian inference, constraining the model's learning of posterior student distributions by ensuring students with similar cognition modes have similar prior distributions. Additionally, to further avoid overfitting, we also use the standard normal distribution as a kind of prior knowledge to constrain model training.

At last, extensive experimental results confirm that CMVF can effectively aid existing KT methods in learning more robust student representations, especially for infrequent students. The contributions of our paper can be summarized as follows:
\begin{itemize}
\item To the best of our knowledge, we are the first KT study to learn robust student representations by building a probabilistic model that generates a distribution for each student.
\item We novelly design a cognition-mode aware prior multinomial distribution to constrain the model to generate similar posterior distributions for students with similar cognition modes which are extracted from their historical practice sequences using a dynamic routing algorithm.
\item CMVF is a general framework that can be incorporated into existing KT methods. Our experimental results demonstrate the superiority and compatibility of the CMVF framework.
\end{itemize}  
 
\section{Related Work}
\textbf{Knowledge Tracing}.  Numerous attempts have been made in KT field over the decades, including probabilistic models \cite{bkt1}, logistic models \cite{irt, afm1}, and deep models \cite{dkt, dkvmn, sakt, akt}. Among the existing methods, most of them revolve around improving the model architecture for modeling students' practice sequences. For example, Bayesian Knowledge Tracing (BKT) \cite{bkt1} uses the Hidden Markova Model (HMM) to trace students' knowledge states. Additive Factor Model (AFM) \cite{afm1} and Performance Factor Analysis (PFA) \cite{pfa} represent students' practices by constructing a practice factor. Inspired of the success of deep learning, Deep Knowledge Tracing (DKT) \cite{dkt} novelly introduces the long short-term memory network (LSTM) \cite{lstm} into KT to represent students' knowledge states with the hidden units. Later, Dynamic Key-Value Memory model (DKVMN) \cite{dkvmn} extends DKT by utilizing a key-value memory network to update students' knowledge states. To better capture long-term practices, SAKT \cite{sakt} introduces the self-attention mechanism into KT field. Convolutional Knowledge Tracing (CKT) \cite{ckt} applies Convolutional Neural Networks (CNN) \cite{cnn} to model students' learning rates. Moreover, there are some works exploring students' sequences in a more granular way. DKT+ \cite{forget} extends DKT to consider forgetting by incorporating multiple types of information related to forgetting. AKT \cite{akt} novelly extends both IRT \cite{irt} and SAKT \cite{sakt} to enhance the interpretability and simulate students' forgetting behaviors. IEKT \cite{iekt} estimates the students’ cognition and assesses knowledge acquisition sensitivity for each record. LPKT \cite{lpkt} monitors students’ knowledge states through directly modeling students’ learning gains and forgetting. LFBKT \cite{lfbkt} introduces the difficulty factor, and models the students’ learning and forgetting behavior according to various influencing factors. DIMKT \cite{dimkt} measures the question difficulty effect and improves KT performance by establishing the relationship between student knowledge states and question difficulty level. However, most of the above methods require enough samples to train model parameters to achieve excellent performance. In reality, the data-sparsity problem is serious since most students practice few times. Although KTM and CL4KT \cite{cl4kt} respectively introduce attribute features and data augmentation methods into the KT field, they still cannot well model the uncertainty of infrequent students. Therefore, in this paper, we propose a general framework CMVF for the KT task to learn robust student representations. 

\textbf{Variational Inference}.  The Variational Auto-Encoder (VAE) \cite{vae} efficiently performs inference and learning by optimizing a latent model using any common estimator. This enables very efficient approximate posterior inference through simple ancestral sampling, allowing for efficient learning of the model parameters without requiring expensive iterative inference schemes (such as MCMC) per datapoint. The learned approximate posterior inference model can be applied in various research fields, including Computer Vision (CV), Natural Language Processing (NLP), and recommendation systems. Despite its success in other fields, VAE has not been utilized in KT to model the uncertainty of infrequent students. 

\textbf{Dynamic Routing Algorithm}.  It is proposed in the Capsule Network \cite{routing}, in which a \emph{capsule} denotes a group of neurons assembled to output a vector. Dynamic routing algorithm is used to learn the weights on the connections between capsules where parameters are optimized by Expectation-Maximization algorithm to overcome several deficiencies and achieves better accuracy. The \emph{capsule} and \emph{dynamic routing mechanism} make the capsule network achieve better performance than conventional deep neural network and achieves advanced performance in multiple research fields \cite{mind, nlp1, nlp2, velf}. 

\section{Preliminaries}
This section provides a brief introduction to the formulation of the KT task and the fundamentals of Variational Inference.

\subsection{Knowledge Tracing Task}
The KT task is a binary classification problem utilizing multi-field factors that may impact a student's learning \cite{dkt, dkvmn}. Given a student's historical practice sequence before time-step $t$, KT aims to predict the probability of the student correctly answer a target question at time $t$. Suppose an online platform collects a dataset $\mathcal{D}$, where each sample $(\boldsymbol{x}, y) \in \mathcal{D}$ denotes a student's practice record. $\boldsymbol{x}$ implies the factors of $\left \{ \emph{student, question, concept, historical practices} \right \}$. $y \in \left\{0, 1 \right\}$ denotes the student's response, where 1 means the answer is correct, and 0 means wrong answer. Let $u$ denote the student ID, $q$ denote the question ID, $c_q$ represent the set of concept IDs related to $q$, and $\chi_u$ denotes the practice sequences. As a result, $\boldsymbol{x}$ can be expressed as $[u, q, c_q,  \chi_u]$.

The widespread use of deep learning in KT has led to the adoption of the \emph{Input $\rightarrow$ Embedding $\rightarrow$ Neural Network $\rightarrow$ Prediction} paradigm in current methods. The \emph{Embedding} layer is a crucial module in KT models, as it compresses high-dimensional ID features into low-dimensional dense representations that are easier to process in subsequent neural network layers, which can be expressed mathematically as follows:
\begin{gather}
\textbf{z}=g_{\phi}(\boldsymbol{x})
\end{gather}
where $\textbf{z}$ denotes the latent embedding space. $g_{\phi}(\cdot)$ represents the mapping function of the embedding layer, and ${\phi}$ represents the parameters of the factor embedding. The resulting embeddings are then concatenated to make the final prediction, which is estimated according to the log-likelihood, as follows:
\begin{gather}
\hat{y}=f_{\theta}(\textbf{z}) \\
\mathcal{L}(\phi, \theta) = -ylog(\hat{y}) - (1-y)log(1-\hat{y})
\end{gather}
where $\theta$ denotes the parameters of \emph{Neural Network}. $\mathcal{L}(\phi, \theta)$ is the optimization objective function in KT \cite{dkt, akt}.

\begin{figure*}[h]
  \centering
  \includegraphics[width=\linewidth]{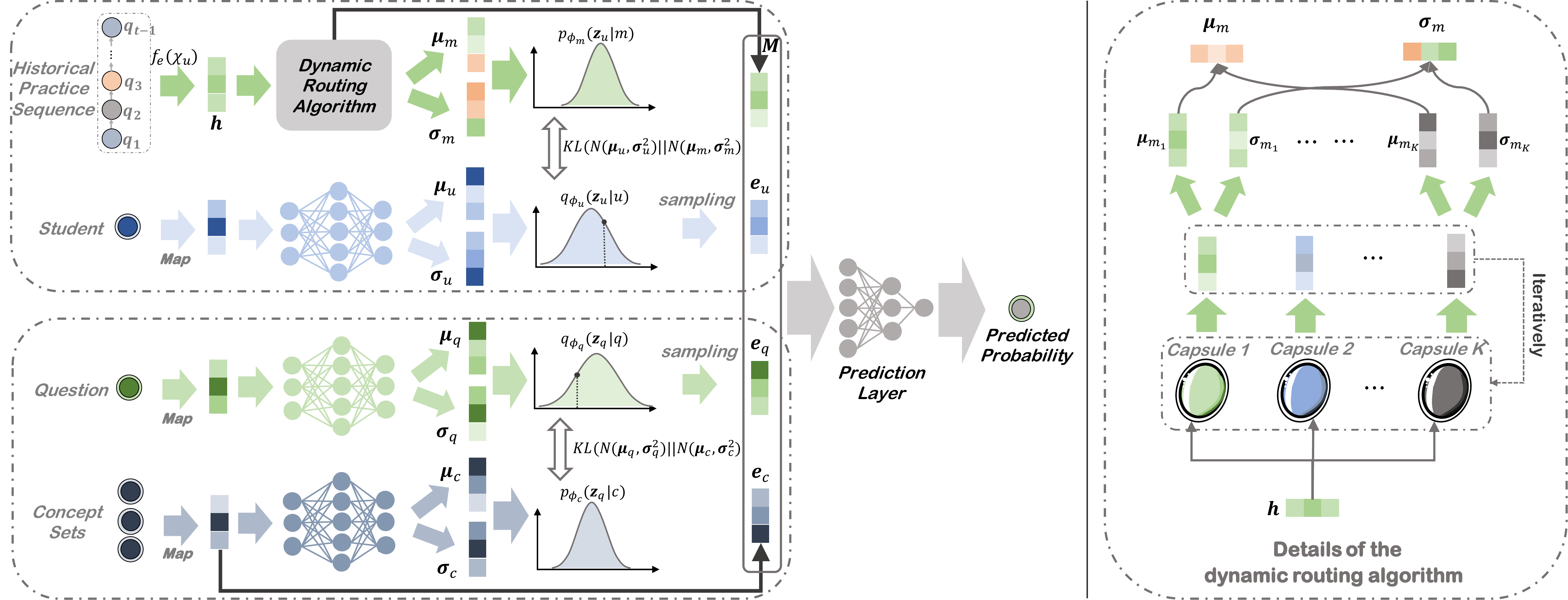}
  \caption{The overall architecture of the Cognition-Mode Aware Variational Representation Learning Framework (CMVF).}
  \label{model}
\end{figure*}

\subsection{Variational Inference}
Bayesian approaches are known for their robustness in handling sparse data. However, they are computationally expensive and may struggle to efficiently learn model parameters using costly iterative inference methods. Variational Auto-Encoder (VAE) \cite{vae} is designed to re-parameterize the variational lower bound, which can be easily optimized using standard gradient descent techniques. To facilitate comprehension of this paper, we present the fundamental assumptions behind VAE. Specifically, we assume that the data can be generated by a latent variable $\boldsymbol{z}$ that conforms to a prior distribution $p(\boldsymbol{z})$. According to Bayes theorem, the integral of the marginal likelihood can be expressed as $p(\boldsymbol{x})=\int p(\boldsymbol{z}) p_{\theta}(\boldsymbol{x} | \boldsymbol{z})$, which is intractable in general. Therefore, VAE approximates $p_{\theta}(\boldsymbol{z} | \boldsymbol{x})$ with a variational approximation $q_{\phi}(\boldsymbol{z} | \boldsymbol{x})$ based on the evidence lower bound (ELBO) as follows: 
\begin{gather}
\begin{aligned}
log \, p(\boldsymbol{x}) &\geq \mathcal{L}({{\phi}}, {\theta};\boldsymbol{x}) \\ 
&=\mathbb{E}_{q_{{{\phi}}}(\boldsymbol{z} | \boldsymbol{x})}[log p_{\theta}(\boldsymbol{x} | \boldsymbol{z})] - KL(q_{{{\phi}}}(\boldsymbol{z} | \boldsymbol{x}) || p(\boldsymbol{z}))
\end{aligned}
 \end{gather}
where $KL$ denotes the Kullback-Leibler divergence. $q_{\phi}(\boldsymbol{z} | \boldsymbol{x})$ and $p_{\theta}(\boldsymbol{x} | \boldsymbol{z})$ are defined as a parameterized function with ${\phi}$ and ${\theta}$, respectively. In this way, variational inference can be easily applied into existing deep KT methods.

\section{Method}
To help KT methods model the uncertainty of the infrequent students, we build a probabilistic representation learning framework to generate a distribution for each student. Simultaneously, we use cognition modes extracted from student practice sequences as a prior distribution $p(\boldsymbol{z})$ for the latent space $\boldsymbol{z}$ in the model, which constrains the distribution estimate in our model. This section provides a detailed description of CMVF, with its structure depicted in Figure \ref{model}.

\subsection{Variational Representation Framework}
Current KT methods utilize the student's representation vector and their historical practice sequences to forecast their responses \cite{ktm}. However, infrequent students with sparse data may exhibit greater uncertainty, resulting in overfitting of the student representation vectors learned by models utilizing only a single point. In order to model the uncertainty of students, estimation of the posterior student distribution over the latent space $\boldsymbol{z}$, denoted as $p(\boldsymbol{z}|\boldsymbol{x})$, is necessary, where the variational inference is applied to reformulate the target of the KT task as $p_{\phi, \theta}(y|\boldsymbol{x}, \boldsymbol{z})$. Due to the intractability of the true posterior distribution, a recognition model $q_{\phi}(\boldsymbol{z}|\boldsymbol{x})$ can be utilized to approximate the true distribution, where $q_{\phi}(\boldsymbol{z}|\boldsymbol{x})$ can be interpreted as a probabilistic encoder generating a distribution (such as a Gaussian distribution) for each student. Based on Bayes theorem, the posterior of $\boldsymbol{z}$ can be represented as $p(\boldsymbol{z}|\boldsymbol{x}) = \frac{p(\boldsymbol{z}, \boldsymbol{x})}{p(\boldsymbol{x})}$. Thus, based on Jensen's inequality, the evidence lower bound (ELBO) of the marginal likelihood $p(\boldsymbol{x})$ can be obtained through the following formula:
\begin{gather}
\begin{aligned}
log \, p(\boldsymbol{x}) &= log \int p(\boldsymbol{x}, \boldsymbol{z}) d \boldsymbol{z}  = log \int p(\boldsymbol{x}, \boldsymbol{z}) \frac{q_{\phi}(\boldsymbol{z}| \boldsymbol{x})}{q_{\phi}(\boldsymbol{z}| \boldsymbol{x})} d \boldsymbol{z} \\
& = log (\mathbb{E}_{q_{\phi}(\boldsymbol{z} | \boldsymbol{x})}[\frac{p(\boldsymbol{x}, \boldsymbol{z})}{q_{\phi}(\boldsymbol{z} | \boldsymbol{x})}]) \\
& \ge \mathbb{E}_{q_{\phi}(\boldsymbol{z} | \boldsymbol{x})} [log p(\boldsymbol{x}, \boldsymbol{z}) - log q_{\phi}(\boldsymbol{z} | \boldsymbol{x})] \\
&= \mathbb{E}_{q_{\phi}(\boldsymbol{z} | \boldsymbol{x})} [log p_{\theta}(\boldsymbol{x}| \boldsymbol{z})] + \mathbb{E}_{q_{\phi}(\boldsymbol{z} | \boldsymbol{x})}[log \frac{p(\boldsymbol{z})}{q_{\phi}(\boldsymbol{z}| \boldsymbol{x})}] \\
&= \mathbb{E}_{q_{\phi}(\boldsymbol{z} | \boldsymbol{x})} [log p_{\theta}(\boldsymbol{x}| \boldsymbol{z})] - KL(q_{\phi}(\boldsymbol{z} | \boldsymbol{x}) || p(\boldsymbol{z}))
\end{aligned}
\end{gather}
where the first term can be regarded as an expected negative reconstruction error (denoted as $\mathcal{L}_{re}$), while the second term serves as a regularizer to constrain the approximate posterior distribution $q_{\phi}(\boldsymbol{z} | \boldsymbol{x})$ via the prior distribution $p(\boldsymbol{z})$. The objective function $\mathcal{L}(\phi, \theta)$ is naturally equivalent to the ELBO, and maximizing $p(\boldsymbol{x})$ is equivalent to minimizing the lower bound. $\phi$ and $\theta$ correspond to the parameters of the latent space and prediction network, respectively.

With the help of the Mean-field Theory \cite{mean}, we assume factors in $\boldsymbol{x}$ are mutually independent and each factor is governed by distinct factors in the variational density. Considering that the question may also suffer from sparsity issues, we also utilize the distribution estimate method to learn the question representation. Consequently, the objective function of our model can be reformulated as follows:
\begin{gather}
\begin{aligned}
\mathcal{L}(\phi, \theta)  = \mathcal{L}_{re} &- KL(q_{{\phi}_u}(\boldsymbol{z}_u | u) || p(\boldsymbol{z}_u)) \\
&- KL(q_{{\phi}_q}(\boldsymbol{z}_q | q) || p(\boldsymbol{z}_q))
\end{aligned}
\end{gather}
where $p(\boldsymbol{z}_u)$ and $p(\boldsymbol{z}_q)$ usually denote the normal Gaussian distribution. However, we suspect that the fixed prior distribution limits the generalization ability of the model due to the significant variation between different students and different questions. As a solution, we propose parameterizing $p(\boldsymbol{z}_q)$ of question as $p_{{\phi}_c}(\boldsymbol{z}_q|c)$, where $c$ denotes the set of concepts related to the question $q$. In this way, questions with similar concepts will have similar prior distributions restrictions. 

Moreover, we propose linking students' cognition modes with their representations to better share global information among students. The student's practice sequence serves not only as a reflection of the knowledge state evolution but also as a supplement to the student's  user portrait. We believe that students display commonalities while practicing questions. For instance, some students may rapidly grasp knowledge concepts and answer questions correctly, which can be interpreted as a high-efficiency cognition mode. To realize this, we suggest extracting cognition modes from students' historical practice sequences and substituting the fixed prior distribution $p(\boldsymbol{z}_u)$ with a cognition-mode aware multinomial distribution $p_{{\phi}_m}(\boldsymbol{z}_u|m)$, where $m$ denotes the cognition mode. This will constrain students with similar cognition modes to have similar prior distributions via KL divergence regularization.

\begin{algorithm} [t]
	\caption{$:$ Dynamic Routing Algorithm.} 
	\label{algorithm1}
	\textbf{Input: } students' historical practice sequence $\chi_{u}$,  iteration times $r$,  the number of capsules $K$. 
	
	encode the student's practice sequence: $\boldsymbol{h} = f_e(\chi_{u})$.
		
	for all capsules: initialize $\boldsymbol{b} \leftarrow \boldsymbol{0}$;
		
		\For {$r$ iterations}
		{
			for the capsule $j$: $w_{j}=softmax(\boldsymbol{b})[j]$ (Eq. 8);
				
			for the capsule $j$: $ \boldsymbol{s}_j = w_{j}\boldsymbol{S}_{j}\boldsymbol{h}$ (Eq. 9);
				
			for the capsule $j$: $  \boldsymbol{m}_j = squash(\boldsymbol{s}_j)$  (Eq. 10);
				
			for the capsule $j$: $b_j \leftarrow b_j + \boldsymbol{m}_j^T \boldsymbol{S}_{j} \boldsymbol{h}$ (Eq. 11);
		 		
		}
		for the capsule $j$: $p(\boldsymbol{m}_j|\chi_u)=\frac{||\boldsymbol{m}_{j}||}{\sum_{k=1}^K ||\boldsymbol{m}_{k}||}$;
		
	\Return $\left\{\boldsymbol{m}_{1}, ..., \boldsymbol{m}_{K}\right\}$, $\left\{ p(\boldsymbol{m}_{1}|\chi_u), ..., p(\boldsymbol{m}_{K}|\chi_u) \right\}$
\end{algorithm}

\textbf{\emph{Extract Cognition Mode. }} Considering the complexity of student learning, we doubt that a single cognition mode would be adequate to accurately characterize a student. To represent students' learning characteristics from multiple perspectives, we utilize the dynamic routing algorithm \cite{mind} to extract cognition modes. Each \emph{capsule} can be interpreted as a cognition mode, as shown in the right half of Figure \ref{model}. Specifically, we first encode the sequence as a multidimensional vector:
\begin{gather}
\boldsymbol{h} = f_e(\chi_{u}), \quad \boldsymbol{h}  \in \mathbb{R}^{d}
\end{gather}
where $f_e(\cdot)$ can be any sequence modeling structures, such as the LSTM structure in DKT \cite{dkt}, and the self-attention structure in AKT \cite{akt}. $d$ is the vector dimension. Then, we initialize a probability vector $\boldsymbol{b}=\boldsymbol{0}\in \mathbb{R}^{K}$, where $K$ is the number of capsules. Each element in $\boldsymbol{b}$ denotes the probability of $\boldsymbol{h}$ belonging to corresponding capsule. The elements in $\boldsymbol{b}$ will be updated by computing the information transferred between the representations of students' sequences and each capsule in an iterative way. In each iteration, the updating process of the capsule $j$ is computed by:
\begin{gather}
w_j = \frac{exp(b_j)}{\sum_{k=1}^K exp(b_k)} \\ 
\boldsymbol{s}_j =  w_j \boldsymbol{S}_j \boldsymbol{h}
\end{gather}
where $w_j$ denotes the probability that $\boldsymbol{h}$ is divided into the capsule $j$. $\boldsymbol{S}_j \in \mathbb{R}^{d \times d}$ denotes the bilinear mapping matrix to be learned. With routing logits calculated, the vector of capsule $j$ can be obtained as $\boldsymbol{s}_j$. The vector-based capsule is expected to be able to represent different properties of an entity, in which the orientation of a capsule represents one property and the length of the capsule is used to represent the probability that the property exists. Hence, a non-linear \emph{squash} function is applied to $\boldsymbol{s}_j$ as follows:
\begin{gather}
 \boldsymbol{m}_j = \frac{||\boldsymbol{s}_j||^2}{1+||\boldsymbol{s}_j||^2} \frac{\boldsymbol{s}_j}{||\boldsymbol{s}_j||}\end{gather}
where the length of the $\boldsymbol{m}_j$ can represent the probability of the current input $\boldsymbol{h}$ belonging to the capsule $j$. The logit $b_j$ will be iteratively refined as follow:
\begin{gather}
b_j =  b_j  + \boldsymbol{m}_j^T \boldsymbol{S}_{j} \boldsymbol{h}
\end{gather}

After $r$ iterations, the final output vector $\boldsymbol{m}_j$ represents the output of the student's sequence $\boldsymbol{h}$ for the cognition mode (i.e., capsule) $j$. To ensure that the probabilities of students belonging to $K$ capsules sum to 1, we perform a softmax calculation for all capsules as follows:
\begin{gather}
p(\boldsymbol{m}_j|\chi_u)=\frac{||\boldsymbol{m}_{j}||}{\sum_{k=1}^K ||\boldsymbol{m}_{k}||}
\end{gather}

The procedure of dynamic routing is summarized in Algorithm \ref{algorithm1}, where $p(\boldsymbol{m}_j|\chi_u)$ denotes the the probability that the student belongs to the $j$-th cognition mode. 

\textbf{\emph{Re-parameterize Trick.}} Next, we apply a re-parameterize trick to generate the posterior distributions as below:
\begin{gather}
q_{{\phi}_u}(\boldsymbol{z}_u | u) = \mathcal{N} (\boldsymbol{\mu}_u, \boldsymbol{\sigma}^2_u) \\
q_{{\phi}_q}(\boldsymbol{z}_q | q) = \mathcal{N} (\boldsymbol{\mu}_q, \boldsymbol{\sigma}^2_q)
\end{gather}
where $\mu_u$ or $\mu_q$, and $\sigma_u$ or $\sigma_q$, are obtained from student ID or question ID with DNNs, as shown in \ref{model}. Similarly, we can obtain the parameterized prior distributions as follows:
\begin{gather}
p_{{\phi}_m}(\boldsymbol{z}_u | m) = \mathcal{N} (\boldsymbol{\mu}_m,  \boldsymbol{\sigma}^2_m) \\
p_{{\phi}_c}(\boldsymbol{z}_q | c) = \mathcal{N} (\boldsymbol{\mu}_c, \boldsymbol{\sigma}^2_c)
\end{gather}
where $\mu_c$ and $\sigma_c$ are obtained from the concept IDs related to the question with DNNs. $\mu_m$ and $\sigma^2_m$ can be calculated as:
\begin{gather}
\boldsymbol{\mu}_m =  \sum_{1 \leq i < K} p(\boldsymbol{m_i}|\chi_u)\boldsymbol{\mu}_{m_i}\\
\boldsymbol{\sigma}^2_m =  \sum_{1 \leq i < K} p(\boldsymbol{m_i}|\chi_u)\boldsymbol{\sigma}^2_{m_i}
\end{gather}
where  $\boldsymbol{\mu}_{m_i}$ and $\boldsymbol{\sigma}^2_{m_i}$ are obtained from the student's $i$-th cognition mode representation with DNNs. Based on the above estimated distributions, we can generate embedding vectors for students and questions by random sampling, as follows:
\begin{gather}
\boldsymbol{e}_u = \boldsymbol{\mu}_u +  \boldsymbol{\sigma}_u \odot \boldsymbol{\epsilon}_u, \quad \boldsymbol{\epsilon}_u \in \mathcal{N} (0,  \boldsymbol{I}) \\
\boldsymbol{e}_q = \boldsymbol{\mu}_q +  \boldsymbol{\sigma}_q \odot \boldsymbol{\epsilon}_q, \quad \boldsymbol{\epsilon}_q \in \mathcal{N} (0,  \boldsymbol{I})
\end{gather}

\subsection{Training Phase}
\subsubsection{Prediction} 
We concatenate the variational embeddings of students and questions to conduct predictions as follows:
\begin{gather}
\hat{y}  = f_{\theta}(\boldsymbol{e}_u, \, \boldsymbol{e}_q, \, \boldsymbol{M}, \, \boldsymbol{e}_c)
\end{gather}
where $\hat{y}$ is the prediction probability value. $\boldsymbol{M}=\sum_{1 \leq i < K} [p(\boldsymbol{m}_{i}|\chi_u) \boldsymbol{m}_{i}]$ denotes the pooling of the student's cognition modes. $\boldsymbol{e}_c$ denotes the mean embedding of concepts related to $q$. $f_{\theta}(\cdot)$ represents the prediction layer, which can be FM \cite{fm}, DNNs, and other structures. The resulting reconstruction error can be calculated by:
\begin{equation}
\mathcal{L}_{re} = -\begin{matrix} \frac{1}{L} \sum_{1 \leq i \leq L} y log \hat{y}_i + (1-y)log(1-\hat{y}_i) \end{matrix}
\end{equation}
where $\hat{y}_i$ denotes the predicted probability of sampled $\epsilon_u$ and $\epsilon_q$. $L$ is the Monte Carlo sampling number for each record.

\subsubsection{Regularized Priors}
Although earlier KT research concurs that student representations can embody student cognition modes, we suspect that representations of active students could be learned from a vast number of observed samples and have the potential to reflect more characteristics apart from the cognition mode. Therefore, though $KL(q_{{\phi}_u}(\boldsymbol{z}_u | u) || p_{{\phi}_m}(\boldsymbol{z}_u|m))$ may be advantageous for modeling infrequent students, there is a risk of impairing the richness of representations for frequent students. Consequently, we propose a personalized prior weight to adaptively adjust the regularization strength of the aforementioned regularization term:
\begin{gather}
 \beta_u = 1 - \frac{1}{1+e^{-n_u}}
\end{gather}
where $n_u$ is the number of students' practice in the training dataset. In this way, the regularization of cognition mode will get weaker as the number of students practice increases, i.e. the $\beta_u$ will get closer to 0. Similarly, we also introduce an individualized prior weight for each questions to help questions that occur frequently in the dataset to learn rich information apart from the information of concepts, i.e. $\beta_q = 1 - \frac{1}{1+e^{-n_q}}$. 

At the same time, if we only depend on the above regularization terms in Eq. 6, our model also tends to be over-fitting. For example, the model may make both $\sigma_{u}$ and $\sigma_{q}$ get close to 0 and only optimize the parameters of $\mu_{u}$ and $\mu_{q}$, which will make the model degrade into normal deep neural networks. Therefore, to avoid the over-fitting issue, we propose to add regularization terms on both distributions of students and questions by forcing the parameterized distributions to be close to a standard normal Gaussian distributions. In this way, the final objective function can be reformulated as below:
\begin{gather}
\begin{aligned}
& \mathcal{L}(\phi, \theta)  = \mathcal{L}_{re}  - \beta_u  KL(q_{{\phi}_u}(\boldsymbol{z}_u | u) || p_{{\phi}_m}(\boldsymbol{z}_u|m))  \\
& - \beta_q  KL(q_{{\phi}_q}(\boldsymbol{z}_q | q) || p_{{\phi}_c}(\boldsymbol{z}_q|c)) \\
&- \alpha [KL(q_{{\phi}_u}(\boldsymbol{z}_u | u) || \mathcal{N} (0,  \boldsymbol{I}))  - KL(q_{{\phi}_q}(\boldsymbol{z}_q | q) || \mathcal{N} (0,  \boldsymbol{I}))] \\
&- \alpha [KL(p_{{\phi}_m}(\boldsymbol{z}_u|m) || \mathcal{N} (0,  \boldsymbol{I}))  - KL(p_{{\phi}_c}(\boldsymbol{z}_q|c) || \mathcal{N} (0,  \boldsymbol{I}))] 
\end{aligned}
\end{gather}
where $\alpha$ allows to achieve a better balance between the latent space independence and reconstruction errors to achieve better prediction performance, as shown in ${\beta}$-VAE \cite{b_vae}. The total training procedure of CMVF is summarized in Algorithm 2.

\subsection{Inferring Phase}
After training CMVF based on the training dataset, we can get the distribution of each student. Thus, during the inference phase, we can regard the mean values of the multinomial distribution as the representation vector of each student or each question to conduct the prediction $\hat{y}$ as follows:
\begin{gather}
\boldsymbol{e}_u = \boldsymbol{\mu}_u, \quad \boldsymbol{e}_q = \boldsymbol{\mu}_q \\
\hat{y}  = f_{\theta}(\boldsymbol{e}_u, \, \boldsymbol{e}_q, \, \boldsymbol{M}, \, \boldsymbol{e}_c)
\end{gather}
where $\boldsymbol{e}_u$ and $\boldsymbol{e}_u$ are obatined by replacing the random sampling vectors $\boldsymbol{\epsilon}_u$ and $\boldsymbol{\epsilon}_q$ in Eq. 19 and Eq. 20 with $\boldsymbol{0}$. 

\begin{algorithm} [t]
	\caption{$:$ Training procedure of CMVF.} 
	\label{alg:algorithm2}
	Initialize parameters $ \theta, \phi$.
	\While {\textnormal{not converged}}{
		\For {mini-batch in training dataset}{
			\For {(t = 0; t $\textless$ T; t = t + 1)}{
				encode the historical practice as $ \boldsymbol{h}$ (Eq. 7);
				
				dynamic routing to model the probability that the input belongs to each cognition mode $p(\boldsymbol{m}_i | \chi_u)$ (Algorithm 1);
				
				randomly samples from the posterior distributions via a re-parametrize trick  (Eq. 13-20);
				
				compute gradient $\nabla_{\theta}\mathcal{L}_(\phi, \theta)$ and $\nabla_{\phi}\mathcal{L}_(\phi, \theta)$ (Eq. 24) ; 
		 		
				}
			
			$ \theta, \phi$ $\leftarrow$ Update parameters using gradients;
		 	}
		 }
	\Return $ \theta, \phi$
\end{algorithm}

\section{Experiments}
In this section, we conduct experiments to evaluate the performance improvement brought by CMVF when plugged into various backbones. There are results can be highlighted:  \\
\textbf{$\bullet$} When plugged into various network backbones, our framework CMVF can achieve the state-of-the-art results compared to various KT baselines on three real-world datasets. \\
\textbf{$\bullet$} CMVF can perform better than both KTM and CL4KT in prediction performance for infrequent students. \\
\textbf{$\bullet$} The multinomial likelihood with dynamic routing favorably over the common Gaussian and logistic likelihoods. 

\begin{table}[t]
	\caption{Statistics of Three Benchmark Datasets.}
	\begin{tabular}{cccc}
    		\hline Dataset & \textbf{ASSIST2012} & \textbf{EdNet} & \textbf{NIPS2020}\\
		\hline \# student & 29,018 & 5,000 & 10,000  \\
		 \# question & 53,091 & 10,779 &27,574  \\
		 \# concept & 265 & 188 & 388   \\
		 \# sample & 2,711,873 & 222,141 & 1,333,879   \\
		  \# avg. concept per question & 1 & 2.21 & 4.17   \\
		 \# sample proportion (infreq.) & 0.88\% & 2.19\% & 6.70\%  \\
		 \# avg. length (infreq.) & 4.11 & 4.87 & 44.65  \\
		\hline
	\end{tabular}
	\label{dataset}
	\vspace{-0.5cm}
\end{table} 

\begin{figure*}[t]
  \centering
  \includegraphics[width=0.9\linewidth]{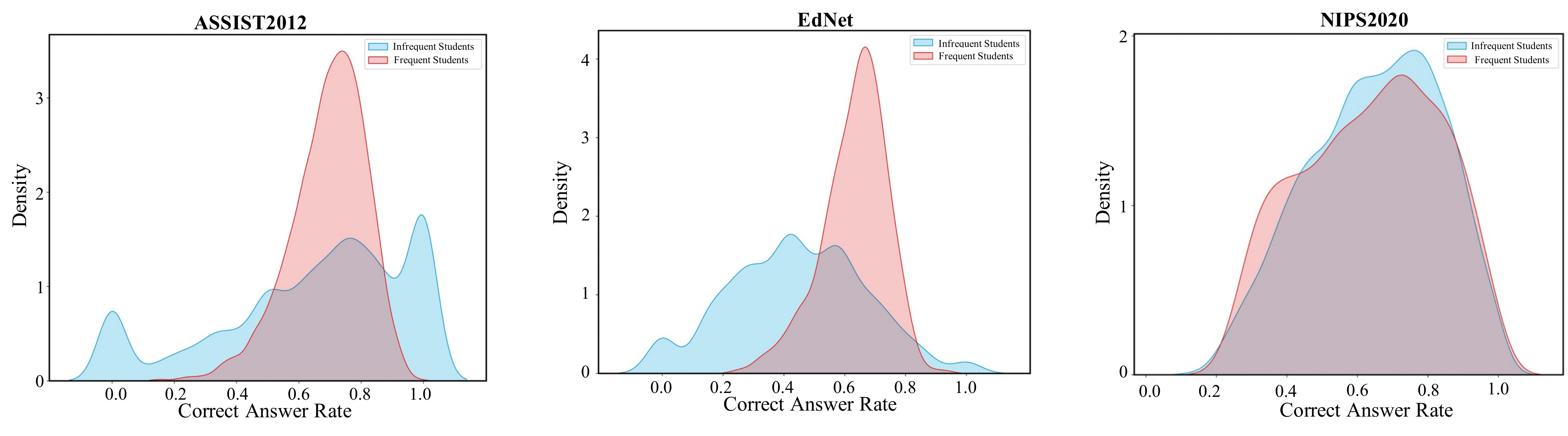}
  \caption{Kernel density estimation of frequent and infrequent students in datasets. Frequent students are those in the top-20\% of practice times.}
  \label{dataset_statistics}
  \vspace{-0.5cm}
\end{figure*}

\subsection{Datasets} 
We evaluate the performance of methods on three datasets: (1) ASSIST2012, (2) EdNet, and (3) NIPS2020. In this paper, we follow \cite{define_ina}, and define infrequent and frequent students as students whose practice times are in the top 80\%-100\% and top 20\%, respectively, according to the reverse order of practice times. The detailed characteristics of three datasets are listed in Table \ref{dataset}, where \emph{infreq} denotes the inactive students.  \\
\textbf{$\bullet$ ASSIST2012 \footnote{https://sites.google.com/site/assistmentsdata/home}}: It is collected from ASSISTments intelligent tutoring system. The average length of practice sequence for infrequent student is 4.11. Attribute features of students include $\emph{student\_class\_id}$, $\emph{teacher\_id}$, $\emph{school\_id}$, $\emph{tutor\_mode}$. \\
\textbf{$\bullet$ Ednet\footnote{https://github.com/riiid/ednet}}: It comes from Santa platform \cite{ednet}and is selected from EdNet-KT1 by previous works \cite{pebg} with 5,000 students' records. Average sequence length for infrequent student is 4.87.  Attribute features are \emph{user\_answer, elapsed\_time}.\\
\textbf{$\bullet$ NIPS2020 \footnote{https://eedi.com/projects/neurips-education-challenge}}: It stems from the NeurIPS 2020 Education Challenge provided by Eedi, with practices from September 2018 to May 2020 \cite{nips2020}. Average sequence length for infrequent student is longer than the other two datasets, 44.65. The attribute features include \emph{Gender, GroupId, SchemeOfWorkId}. 

Figure \ref{dataset_statistics} displays the distribution of frequent and infrequent students. In ASSIST2012 and EdNet datasets, the correct answer rate for frequent students is generally more concentrated, while the distribution for infrequent students is more scattered, indicating higher uncertainty. This finding verifies the necessity of modeling uncertainty. In the NIPS2020 dataset, our definition of infrequent students resulted in an average sequence length of 44.65, as shown in Table \ref{dataset}, leading to a similar distribution of infrequent students to frequent students.

\subsection{Baselines}
\noindent \textbf{$\bullet$ DKT} \cite{dkt}: This method is the first to apply deep neural network into the KT field. It utilizes LSTMs to model the evolution of students' knowledge states with the hidden units.\\
\textbf{$\bullet$ DKVMN} \cite{dkvmn}: This method extends DKT by using a key-value memory network to store and update students' knowledge states based on a static key matrix and a dynamic value matrix and can mine the relationship between concepts.\\
\textbf{$\bullet$ KTM} \cite{ktm}: This method is the first to apply the FM \cite{fm} into modeling multiple factors related to students' learning and is a quite generic framework for factor analysis methods.\\
\textbf{$\bullet$ SAKT} \cite{sakt}: This method applies the structure of Transformer to model students' practice sequences to enhance the ability to capture long-term dependencies between practices. \\
\textbf{$\bullet$ AKT} \cite{akt}: This method proposes to update the practice representation with awareness of contexts and introduces the IRT \cite{irt} to acquire question embeddings. \\
\textbf{$\bullet$ DIMKT} \cite{dimkt}: This method measures students' subjective feelings of question difficulty and estimates students' knowledge acquisition while answering questions of different difficulty levels to model the evolution of knowledge states.\\
\textbf{$\bullet$ KTM+} \cite{ktm}: KTM can freely incorporate attribute features into predicting students' responses. In this paper, we define the KTM+ as the KTM-based method with attribute features. \\ 
\textbf{$\bullet$ CL4KT} \cite{cl4kt}: This method introduces a contrastive learning framework that reveals semantically similar or dissimilar examples of a learning history with data augmentation methods. 

To assess the effectiveness of CMVF, we select multiple classic methods including DKT (the cornerstone of deep KT methods), KTM (a general framework for factor analysis methods), and DIMKT (a state-of-the-art method) as the network backbone for the CMVF framework. \\
\textbf{$\bullet$ CMVF+DKT}: We select the hidden state of LSTM in DKT, i.e. $\boldsymbol{h}$, as the input of the dynamic routing algorithm to extract cognition mode $\boldsymbol{M}$. To incorporate student embedding into the DKT, we apply a fully-connected layer for final prediction, i.e. $y = \sigma(MLP(\boldsymbol{M} \oplus \boldsymbol{e}_u \oplus \boldsymbol{e}_{q}))$, where $\boldsymbol{e}_u$ and  $\boldsymbol{e}_q$ denote the variational embeddings of students and questions.  $\sigma(\cdot)$ denotes the sigmoid function. \\
\textbf{$\bullet$ CMVF+KTM}: We exploit \emph{wins} and \emph{fails} factors in KTM \cite{ktm} to extract the cognition modes, and replace the original embeddings of factors with variational embeddings. Moreover, we replace the representations of \emph{wins} and \emph{fails} factors in FM with the students' cognition mode representation. \\
\textbf{$\bullet$ CMVF+DIMKT}: The output of \emph{knowledge state updating} in DIMKT is utilized to extract cognition modes $\boldsymbol{M}$. As the original DIMKT did not incorporate student representations, we fuse difficulty-enhanced question embedding and student embedding into $\boldsymbol{x}=\boldsymbol{W}^T [\boldsymbol{e}_u \oplus \boldsymbol{e}_q \oplus \boldsymbol{e}_c \oplus \boldsymbol{QS} \oplus \boldsymbol{KC}] + \boldsymbol{b}$, where $\boldsymbol{QS}$ and $\boldsymbol{KC}$ denote the difficulty of question and concept, respectively. At last, we make the prediction as $y = \sigma(\boldsymbol{M}\cdot \boldsymbol{x})$. 

\begin{table*}[t]
	\caption{The performance on three benchmark datasets of KT methods.  * indicates the improvement is statistically significant at the significance level of 0.05 over the best baseline on AUC. \emph{Infreq} denotes the infrequent student groups.}
	\resizebox{\textwidth}{!}{
	\begin{tabular}{c|c||c||cccccc||cc||ccc}
		\hline \hline 
		\multirow{2}*{Dataset} & \multirow{2}*{\makecell[c]{Test \\ Sets}} & \multirow{2}*{Metric} & \multirow{2}*{DKT} & \multirow{2}*{DKVMN} & \multirow{2}*{KTM} & \multirow{2}*{SAKT} & \multirow{2}*{AKT} & \multirow{2}*{DIMKT} & \multirow{2}*{KTM+} & \multirow{2}*{CL4KT} & \multirow{2}*{\makecell[c]{CMVF +\\ DKT}} & \multirow{2}*{\makecell[c]{CMVF+ \\ KTM}} & \multirow{2}*{\makecell[c]{CMVF + \\ DIMKT}} \\
		& & & & & && & & & & & & \\ 
		\hline \hline
		\multirow{6}*{ASSIST2012} & \multirow{3}*{Overall} &ACC &0.7366 &0.7348&0.7291&0.7424 &0.7452 &0.7525 &0.7362 &0.7501 &0.7530 &0.7513 &\textbf{0.7605*} \\ 
		& &AUC &0.7410 &0.7399 &0.7355 &0.7538&0.7593 &0.7815 &0.7495 &0.7663 &0.7757 &0.7716 &\textbf{0.7946*} \\  
		& &\emph{RealImpr}&0.0\% &-0.5\% &-2.3\% & 5.3\%& 7.6\% &16.8\% &3.5\% &10.5\% &14.4\% &12.7\% & \textbf{22.2\%} \\ 
		\cline{2-14}
		& \multirow{3}*{Infreq} &ACC &0.7025 &0.7002&0.7118 &0.7043 &0.7224 &0.7296 &0.7143 &0.7281 &0.7446 & 0.7286&\textbf{0.7512*} \\ 
		& &AUC &0.7126 &0.7093 &0.7008 &0.7180&0.7356 &0.7388 &0.7059 &0.7412 &0.7696 &0.7546 &\textbf{0.7875*} \\ 
		& &\emph{RealImpr}&0.0\% &-1.6\% &-5.6\% & 2.5\%& 10.8\% &12.3\% &-3.2\% &13.5\% &26.8\% &19.8\% & \textbf{35.2\%}\\ 
		\hline \hline 
		\multirow{6}*{EdNet} & \multirow{3}*{Overall} &ACC &0.6785 &0.6816&0.6714 &0.6774 &0.6909 &0.6983 &0.6758 &0.6948 &0.6764 &0.6890 &\textbf{0.7043*}  \\ 
		& &AUC &0.7243 & 0.7214&0.7193 &0.7220&0.7300 &0.7391 &0.7269 &0.7323 &0.7411 &0.7499 &\textbf{0.7593*} \\  
		& &\emph{RealImpr}&0.0\% &-1.3\% &-2.2\% & -1.0\%&2.5\% &6.6\% &1.2\% &3.6\% &7.5\% &11.4\% &\textbf{15.6\%} \\ 
		\cline{2-14}
		& \multirow{3}*{Infreq} &ACC &0.5948 &0.5993&0.5860 &0.6019 &0.6003 &0.6059 &0.5902 &0.6031 &0.6201 &0.6106 &\textbf{0.6254*} \\ 
		& &AUC &0.6139 &0.6172 &0.6078 &0.6316&0.6349 &0.6437 &0.6104 &0.6403 &0.6683 &0.6675 & \textbf{0.6751*} \\ 
		& &\emph{RealImpr} &0.0\% &2.9\% &-5.4\% &15.5\% &18.4\% &26.2\% &-3.1\% &23.2\% &47.8\% &47.1\% & \textbf{53.7\%} \\ 
		\hline \hline 
		\multirow{6}*{NIPS2020} & \multirow{3}*{Overall} &ACC &0.7276 &0.7258&0.7218 &0.7259 &0.7280 &0.7337 &0.7258 &0.7328 &0.7320 & 0.7256 &\textbf{0.7392*} \\ 
		& &AUC &0.7866 &0.7821 &0.7770 &0.7831&0.7878 &0.7979 &0.7861 &0.7980 &0.8002 & 0.7917 &\textbf{0.8044*} \\  
		& &\emph{RealImpr}&0.0\% &-1.6\% &-3.3\% &-1.2\%&0.4\% &3.9\% &-0.2\% &4.0\% &4.7\% &1.8\% &\textbf{6.2\%} \\ 
		\cline{2-14}
		& \multirow{3}*{Infreq} &ACC &0.7098 &0.7086&0.7016 &0.7130 &0.7194 &0.7214 &0.7056 &0.7232 &0.7213 &0.7161 &\textbf{0.7299*} \\ 
		& &AUC &0.7637 &0.7638 &0.7569 &0.7651&0.7710 &0.7780 &0.7609 &0.7813 &0.7864 &0.7773 &\textbf{0.7921*} \\ 
		& &\emph{RealImpr}&0.0\% &0.0\% &-2.6\% &0.5\%&2.8\% &5.4\% &-1.1\% &6.7\% &8.6\% &5.2\% &\textbf{10.8\%} \\ 
		\hline \hline 
	\end{tabular}
	}
	\label{performance}
	\vspace{-0.3cm}
\end{table*}  

\subsection{Experimental Settings} 
\subsubsection{Implementation Details} 
Unlike previous works that typically discard students with less than 10 practice behavior records \cite{cl4kt, dimkt}, we retain more infrequent students by only discarding those with fewer than 3 records. For the three datasets, we use the top 80\% of each student's records as the training set to predict their remaining behavior, running five experiments with varying random seeds for statistical significance analysis. We set the embedding size to 64 and truncate student sequences longer than 200. Mini-batch Adam \cite{adam} optimization is used for all methods, with a mini-batch size of 2048, where the learning rate is searched from $\left\{1e-5, 5e-4, 1e-4, ..., 1e-2 \right\}$. The Xavier parameter initialization method \cite{initializer} is used to initialize parameters. All models are trained on a Linux server with two Intel(R) Xeon(R) CPU E5-2620 v4 and a NVIDIA TITAN XP GPU.

\begin{figure*}[t]
  \centering
  \includegraphics[width=0.95\linewidth]{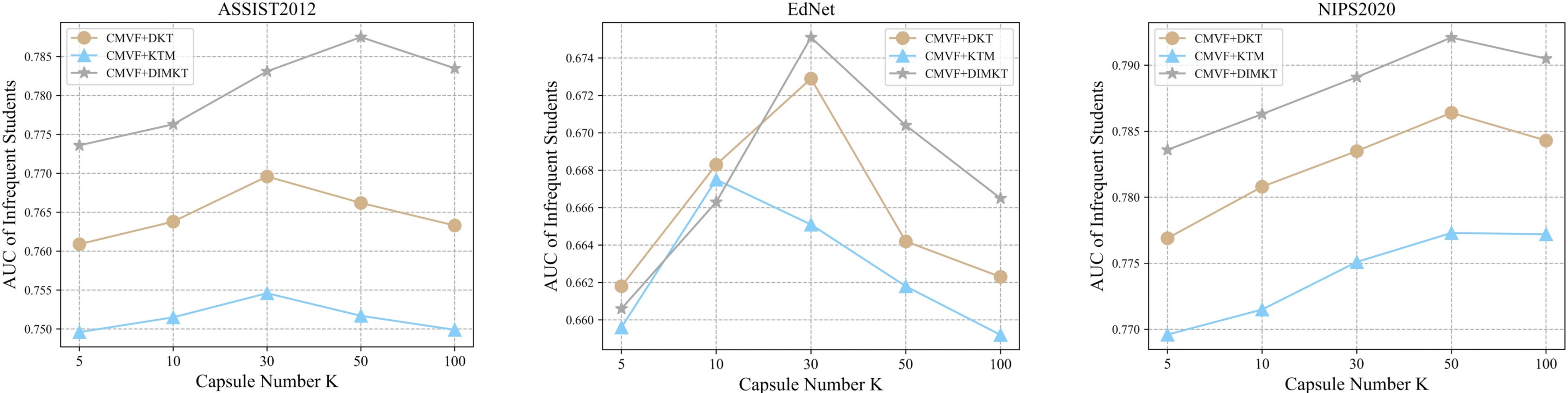}
  \caption{The sensitivity analysis of the number of capsules $K$ in the three datasets.}
  \label{auc_capsule}
  \vspace{-0.3cm}
\end{figure*}

\begin{figure*}[t]
  \centering
  \includegraphics[width=0.95\linewidth]{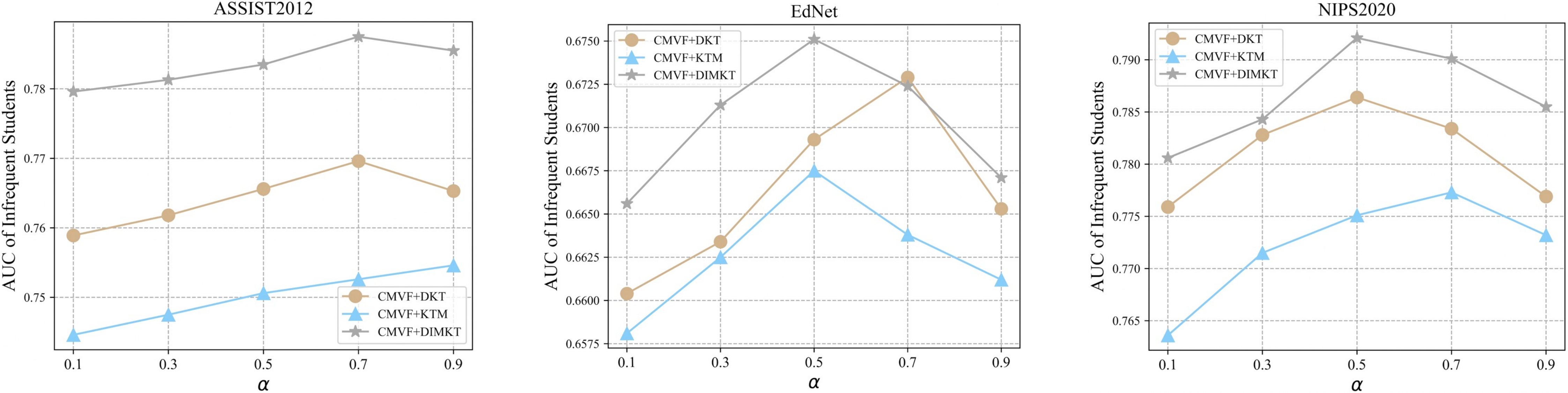}
  \caption{The sensitivity analysis of the regularization coefficient $\alpha$ in the three datasets.}
  \label{auc_alpha}
  \vspace{-0.3cm}
\end{figure*}

\subsubsection{Evaluation Metrics}
Accuracy (ACC) and Area Under Curve (AUC) are two common metrics for the KT task, where the threshold of ACC is set as 0.5. AUC is more robust than ACC, which aims to measure the rank performance of methods and is in line with the application requirements of KT to recommend questions to students. Therefore, AUC is the main metric in this paper. In addition, following \cite{velf}, we introduce the \emph{RealImpr} metric to measure the relative improvement over the base method, which can be defined as follows:
\begin{gather}
RealImpr = (\frac{AUC(target)-0.5}{AUC(base)-0.5}-1) \times 100\%
\end{gather}
where we adopt the classical DKT method as the base model.

\subsection{Performance on Predicting Students' Responses}
This section presents a comparison of CMVF with state-of-the-art methods using different backbones, with Table \ref{performance} showing the mean results of three metrics for all methods on the three datasets, with the highest score highlighted in bold.

Based on Table \ref{performance}, we observe that CMVF significantly improves the accuracy of KT methods in predicting infrequent student groups, demonstrating its reliability in modeling uncertainty for these students. We also find a more significant improvement in CMVF's scored AUC for infrequent student groups compared to the overall AUC improvement, supporting our hypothesis that CMVF has a greater effect on infrequent students who practice less and exhibit greater uncertainty. However, since the proportion of infrequent students in the entire dataset is relatively small, the overall AUC improvement is relatively lower than that for infrequent students.

From Table \ref{performance}, we also observe that most KT models perform poorly on infrequent student groups, particularly in the EdNet dataset. As shown in Table \ref{dataset}, the average practice sequence length of infrequent students in EdNet is quite short, only 4.87. Meanwhile, as shown in Figure \ref{dataset_statistics}, the distribution of frequent and infrequent students in EdNet is significantly different from that of the other two datasets. However, due to unbalanced training sets and a large proportion of frequent student samples, model parameter learning is often influenced more by frequent students, leading to overwhelmed personalization of infrequent students and difficulty in distinguishing the difference between two significantly different distributions. 

Furthermore, Table \ref{performance} shows that introducing attribute features (KTM+) and data augmentation methods (CL4KT) outperforms KTM and AKT (the backbone of CL4KT) in predicting infrequent students. The method of data augmentation combined with contrastive learning is found to be better than introducing attribute features in improving the model's ability to predict infrequent students, as shown in Table \ref{performance}. In addition, CL4KT improves the performance of AKT more significantly in NIPS2020, where the average length of student sequences is longer, allowing for more reliable enhanced samples. However, the limited benefit of CL4KT on AKT in both EdNet and ASSIST2012 datasets, where the average sequences are shorter, highlights the limitations of data augmentation methods. To further compare the superiority of CMVF with the above two methods, we apply these methods uniformly to the DIMKT backbone for experiments, as shown in Table  \ref{augmen}. The results demonstrate that CMVF significantly outperforms the other two methods in all three datasets.

\subsection{Sensitivity Analysis of Hyper-Parameters}
In this paper, CMVF includes two manually-tuned hyper-parameters: the number of capsules $K$ and the coefficient $\alpha$ of the regularization term. Therefore, in this section, we conducted experiments with different values of these parameters to investigate their impact on model performance.

\subsubsection{Sensitivity Analysis of Capsule Number}
We evaluate CMVF's performance using five different capsule numbers with three backbone models (DKT, KTM, and DIMKT), namely $K=5, 10, 30, 50$, and $100$, as shown in Figure \ref{auc_capsule}. Small values of K (e.g., 5) are found to lead to low prediction performance because the number of capsules determines the angle of mining students' cognition modes. The more angles, the richer the information that can be captured in the students' practice sequences. However, if $K$ is too large, the model may overfit due to a lack of data for optimizing parameters. For convenience, we should choose a value between 30 and 50 for the number of capsules that achieves good performance and is common to most datasets, as illustrated in Figure \ref{auc_capsule}.

\subsubsection{Sensitivity Analysis of Regularization Coefficient}
Since the regularization term of the standard normal Gaussian distribution should be an auxiliary task and the main task is still the reconstruction error, we set the coefficient a of the regularization term in the range of 0 to 1 and evaluate CMVF's performance with five different values: 0.1, 0.3, 0.5, 0.7, and 0.9. The experimental results on the three datasets are shown in Figure \ref{auc_alpha}. The KT task is intended to predict student responses to practice, not to maximize the probability likelihood of the simulated student distribution. Therefore, the closer the value of $\alpha$ is to 1, the more likely our task will deviate from the original intention, leading to decreased prediction performance of the model for responses. On the other hand, too small a value may cause our regularization to lose its effect and degenerate into an ordinary KT model with a deep neural network, leading to overfitting. Therefore, setting the value of $\alpha$ in the range of 0.5-0.7 can achieve a more universal effect.

\begin{table}[t]
	\caption{The performance of DIMKT on infrequent students with different methods on three datasets.  $std \le 0.1\%$. }
	\resizebox{\linewidth}{!}{
	\begin{tabular}{c|cc|cc|cc}
		\hline \hline 
		\multirow{2}*{Method} & \multicolumn{2}{c|}{ASSIST2012}& \multicolumn{2}{c|}{EdNet}& \multicolumn{2}{c}{NIPS2020}\\ 
		& ACC & AUC & ACC & AUC & ACC & AUC \\ 
		\hline \hline 
		AF+DIMKT &0.7331 &0.7453 &0.6083 &0.6479 &0.7231 &0.7814\\
		CL4+DIMKT&0.7356 &0.7498 &0.6103 &0.6502 &0.7245 &0.7846\\
		CMVF+DIMKT& \textbf{0.7512}& \textbf{0.7875}& \textbf{0.6254}& \textbf{0.6751}& \textbf{0.7299}& \textbf{0.7921}\\
		\hline \hline 
	\end{tabular}
	}
	\label{augmen}
	\vspace{-0.3cm}
\end{table} 

\begin{table}[t]
	\caption{The performance of DKT on infrequent students with different variants of CMVF on three datasets.  $std \le 0.1\%$. }
	\resizebox{\linewidth}{!}{
	\begin{tabular}{c|cc|cc|cc}
		\hline \hline 
		\multirow{2}*{Method} & \multicolumn{2}{c|}{ASSIST2012}& \multicolumn{2}{c|}{EdNet}& \multicolumn{2}{c}{NIPS2020}\\ 
		& ACC & AUC & ACC & AUC & ACC & AUC \\ 
		\hline \hline 
		CMVF & \textbf{0.7446}& \textbf{0.7696} & \textbf{0.6201} & \textbf{0.6683} & \textbf{0.7213} & \textbf{0.7864}\\
		CMVF(Uniform) &0.7418 &0.7563 &0.6054 &0.6488 &0.7124 &0.7720\\
		CMVF(R-Capsule) &0.7413 &0.7582 &0.6116 &0.6585 &0.7135 &0.7785\\
		CMVF(R-Reg) &0.7442 &0.7610 &0.6094 &0.6572 &0.7145 &0.7782\\
		CMVF(Point) &0.7394 &0.7453 &0.6001 &0.6209 &0.7117 &0.7689\\
		\hline \hline 
	\end{tabular}
	}
	\label{ablation}
	\vspace{-0.5cm}
\end{table} 

\subsection{Ablation Study}
To get deep insights into the contributions of components in CMVF, we conduct ablation studies by applying multiple variants to the DKT model which is one of the lightest structures. The details of four variants are as follows: \\
\textbf{$\bullet$ CMVF(Uniform)}. We replace the prior distribution of CMPF with the uniform prior distributions as VAE does \cite{vae}. \\
\textbf{$\bullet$ CMVF(R-Capsule)}. We remove the dynamic routing component and directly input the representations of students' practices into re-parametrize component to regularize training. \\
\textbf{$\bullet$ CMVF(R-Reg)}. We remove the two mutual regularization terms in CMVF, i.e. $KL(q_{{\phi}_u}(\boldsymbol{z}_u | u) || p_{{\phi}_m}(\boldsymbol{z}_u|m))$ and $KL(q_{{\phi}_q}(\boldsymbol{z}_q | q) || p_{{\phi}_c}(\boldsymbol{z}_q|c))$, and only keep regularization terms of  standard normal Gaussian distribution. \\
\textbf{$\bullet$ CMVF(Point)}. We replace the distinct distribution estimate with the static representation estimate by directly adopting the mean values $\boldsymbol{\mu}_u$ and $\boldsymbol{\mu}_q$ as the embeddings.

Table \ref{ablation} reports the average results of five experiments. Firstly, we observe that CMVF(Uniform) and CMVF(R-Reg) exhibit similar performance, but are inferior to CMVF. This highlights the benefit of using a parameterized prior distribution for perceived cognition modes of students, which prevents infrequent students from being overwhelmed by individuation. Secondly, CMVF is more effective than CMVF(R-Capsule), because the dynamic routing algorithm can extract student practice information from multiple capsules, enhancing student sequence modeling. Furthermore, CMVF(R-Capsule) outperforms both CMVF(Uniform) and CMVF(R-Reg), indicating that even if the dynamic routing algorithm is removed, performance can still improve as long as student practice information is considered, highlighting the benefit of parameterized prior.  Finally, CMVF(Point) performs the worst among the four variants, providing strong evidence that distribution estimates outperform point estimates.

\section{Conclusions}
This paper proposes a general representation learning framework for KT called Cognition-Mode Aware Variational Framework (CMVF) to learn robust student representations based on variational inference and the dynamic routing algorithm. Specifically, to better model the uncertainty of students with few practice records, we introduce a probabilistic model to generate a distribution for each student and use Bayesian inference for parameter estimation. Meanwhile, to better help the model estimate the distribution of infrequent students, we extract cognition modes from students' historical practice sequences through dynamic routing to set similar prior distributions for students with similar characteristics, thereby pulling distributions of students with similar cognition modes closer. Extensive experimental results validate the superiority and compatibility of our framework CMVF. 

\section*{Acknowledgment}
This work is supported by 2022 Beijing Higher Education ``Undergraduate Teaching Reform and Innovation Project'' and 2022 Education and Teaching Reform Project of Beijing University of Posts and Telecommunications (2022JXYJ-F01).

\end{document}